\documentclass[11pt,a4paper]{article}
\usepackage[hyperref]{acl2020}
\usepackage{times}
\usepackage{latexsym}

\usepackage{booktabs}
\usepackage{tabularx}
\usepackage{todonotes}
\usepackage{multirow}
\usepackage{url}
\usepackage{amsmath}
\usepackage{amssymb}
\usepackage{tablefootnote}
\usepackage{caption}
\usepackage{subcaption}
\usepackage{bbm}
\usepackage{comment}
\usepackage{xspace,mfirstuc,tabulary}
\newcolumntype{Y}{>{\centering\arraybackslash}X}
\usepackage{mdframed}

\aclfinalcopy 

\newcommand{\streg}{\textsc{StructuredRegex}}
\newcommand{\dpreg}{\textsc{DeepRegex}}
\newcommand{\dpfilter}{\textsc{DeepRegex+Filter}}
\newcommand{\kb}{\textsc{KB13}}
\newcommand{\turk}{\textsc{NL-Turk}}
\newcommand{\so}{\textsc{StackOverflow}}

\newcommand{\regdsl}[1]{\texttt{\small #1}}

\newcommand{\gramrule}[2]{{\regdsl{#1} $\rightarrow$ \regdsl{#2}}}
\newcommand{\gramdiv}{$|$}
\title{Benchmarking Multimodal Regex Synthesis with Complex Structures}

\author{Xi Ye\quad Qiaochu Chen\quad Isil Dillig\quad Greg Durrett\\
  Department of Computer Science \\
  The University of Texas at Austin \\
  \regdsl{\{xiye,qchen,isil,gdurrett\}@cs.utexas.edu} \\ }

\date{}

\begin{document}
\maketitle
\begin{abstract}
Existing datasets for regular expression (regex) generation from natural language are limited in complexity; compared to regex tasks that users post on StackOverflow, the regexes in these datasets are simple, and the language used to describe them is not diverse. We introduce \streg{}, a new regex synthesis dataset differing from prior ones in three aspects.
First, to obtain structurally complex and realistic regexes, we generate the regexes using a probabilistic grammar with pre-defined macros observed from real-world StackOverflow posts. Second, to obtain linguistically diverse natural language descriptions, we show crowdworkers abstract depictions of the underlying regex and ask them to describe the pattern they see, rather than having them paraphrase synthetic language. Third, we augment each regex example with a collection of strings that are and are not matched by the ground truth regex, similar to how real users give examples. Our quantitative and qualitative analysis demonstrates the advantages of \streg{} over prior datasets. Further experimental results using various multimodal synthesis techniques highlight the challenge presented by our dataset, including non-local constraints and multi-modal inputs.\footnote{Code and data available at \url{https://www.cs.utexas.edu/~xiye/streg/}.}   

\end{abstract}

\begin{figure}[h]
\centering
\includegraphics[width=\linewidth, trim=450 20 450 20,clip]{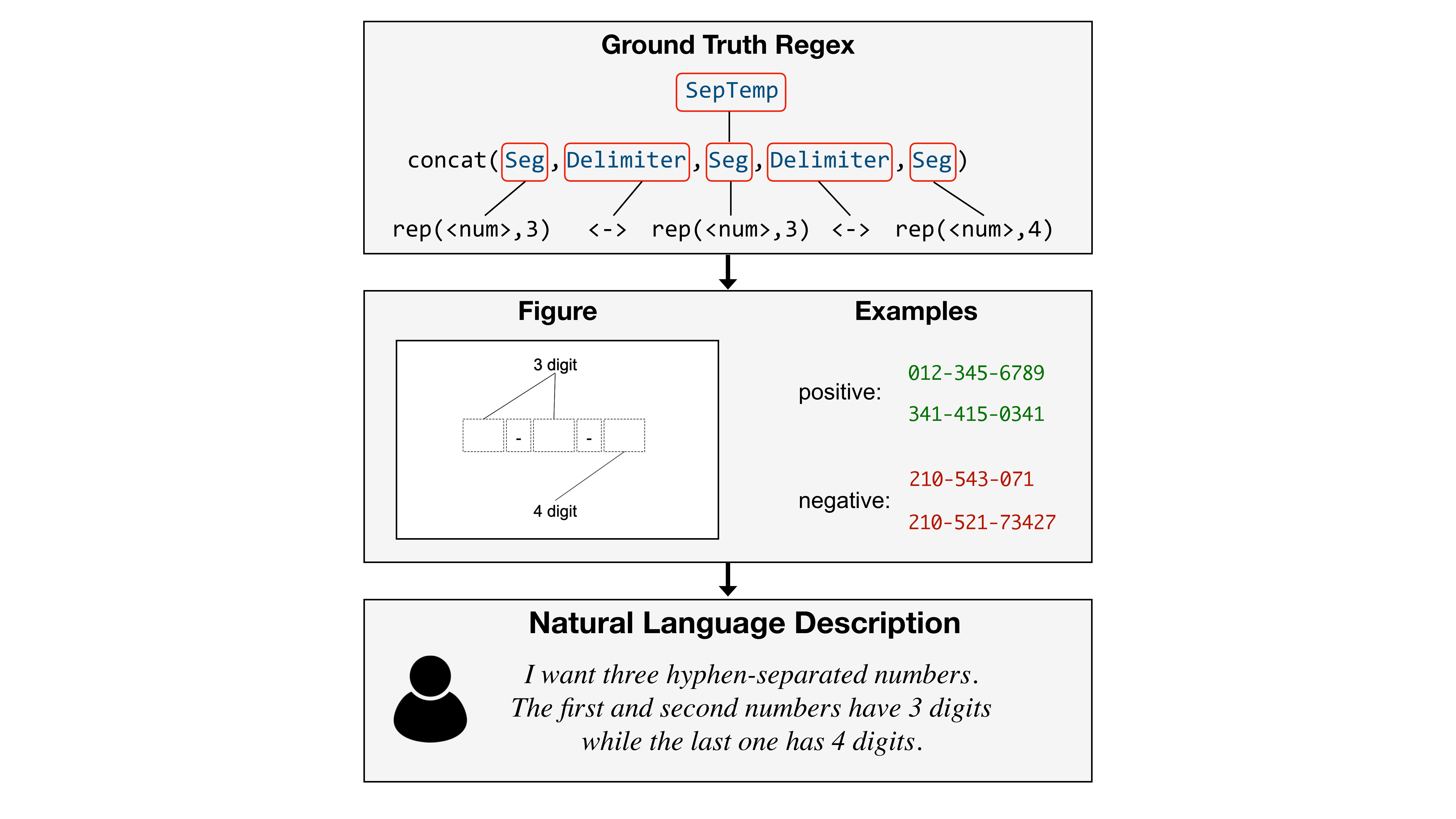}
\centering
\caption{Our dataset collection process. A regex is sampled from our grammar, then we render an abstract figure and generate distinguishing positive/negative examples. We present the figure and examples to crowdworkers to collect natural language descriptions.}
\label{fig:overall_proc}
\end{figure}

\section{Introduction}

\begin{figure*}[h!]
    \small
    \begin{tabularx}{\linewidth}{l X}
    \toprule
    \multirow{2}{*}{(a)}  & \textit{I need to validate the next pattern: starts with ``C0'' and finish with 4 digits exactly.} \\
        & \regdsl{and(startwith(<C0>)),endwith(rep(<num>,4)))} \\

    \midrule
        \multirow{2}{*}{(b)} & \textit{i need regular expression for : one or two digits then "." and one or two digits.} \\
        & \regdsl{concat(reprange(<num>,1,2),concat(<.>,reprange(<num>,1,2)))} \\
    \midrule
         \multirow{2}{*}{(c)} &\textit{The input will be in the form a colon (:) separated tuple of three values. The first value will be an integer, with the other two values being either numeric or a string.} \\
        & \regdsl{concat(repatleast(<num>,1),rep(concat(<:>,or(repatleast(<let>,1), repatleast(<num>,1))),2))} \\
    \bottomrule
    \end{tabularx}
    \caption{Examples of complex regexes from \so{}. Each regex can be viewed as a set of components composed with a high-level template. Regex (a), for example, can be as viewed the intersection of two constraints specifying the characteristics of the desired regex. (\regdsl{rep} means repeat).} 
    \label{so_comp_reg}
\end{figure*}

Regular expressions (regexes) are known for their usefulness and wide applicability, and yet they are hard to understand and write, even for many programmers \cite{fried06}. Recent research has therefore studied how to construct regexes from natural language (NL) descriptions, leading to the emergence of NL-to-regex datasets including \kb{} \cite{KB13} and \turk{} \cite{deepregex}. However, \kb{} is small in size, with only 814 NL-regex pairs with even fewer distinct regexes. \citet{deepregex} subsequently employed a generate-and-paraphrase procedure \cite{overnight} to create the larger \turk{} dataset. However, the regexes in this dataset are very simple, and the descriptions are short, formulaic, and not linguistically diverse because of the paraphrasing annotation procedure \cite{herzig19}. As a result, even when models achieve credible performance on these datasets, they completely fail when evaluated on the \so{} dataset \cite{deepsketch}, a real-world dataset collected from users seeking help on StackOverflow. The limited size of this dataset (only 62 NL-regex pairs) makes it unsuitable for large-scale training, and critically, the complexity of regexes it features means that regex synthesis systems must leverage the user-provided positive and negative examples (strings that should be matched or rejected by the target regex) in order to do well.

To enable the development of large-scale neural models in this more realistic regex setting, we present \streg{}, a new dataset of English language descriptions and positive/negative examples associated with complex regexes. Using a new data collection procedure (Figure~\ref{fig:overall_proc}), our dataset addresses two major limitations in \turk{}. First, we generate our regexes using a structured probabilistic grammar which includes \emph{macro} rules defining high-level templates and constructions that involve multiple basic operators. These grammar structures allow us to sample more realistic regexes, with more terminals and operators, while avoiding vacuous regexes. By contrast, the random sampling procedure in \turk{} leads to simple regexes, and attempting to sample more complex regexes results in atypical regex structures or even contradictory regexes that do not match any string values \cite{deepsketch}. Second, to achieve more realistic language descriptions, we prompt Turkers to write descriptions based on abstract figures that show the desired regexes. We design a set of visual symbols and glyphs to \emph{draw} a given regex with minimal textual hints. We thereby avoid priming Turkers to a particular way of describing things, hence yielding more linguistically diverse descriptions.

Using this methodology, we collect a total of 3,520 English descriptions, paired with ground truth regexes and associated positive/negative examples. We conduct a comprehensive analysis and demonstrate several linguistic features present in our dataset which do not occur in past datasets. We evaluate a set of baselines, including grammar-based methods and neural models, on our dataset. In addition, we propose a novel decoding algorithm that integrates constrained decoding using positive/negative examples during inference: this demonstrates the potential of our dataset to enable work at the intersection of NLP and program synthesis. The performance of the best existing approach on \streg{} only reaches 37\%, which is far behind 84\% on \turk{}. However, this simple model can nevertheless solve 13\% of the \so{} dataset, indicating that further progress on this dataset can be useful for real-world scenarios.

\begin{figure*}
\small
  {
  \centering
\includegraphics[width=\linewidth, trim=130 235 110 235,clip]{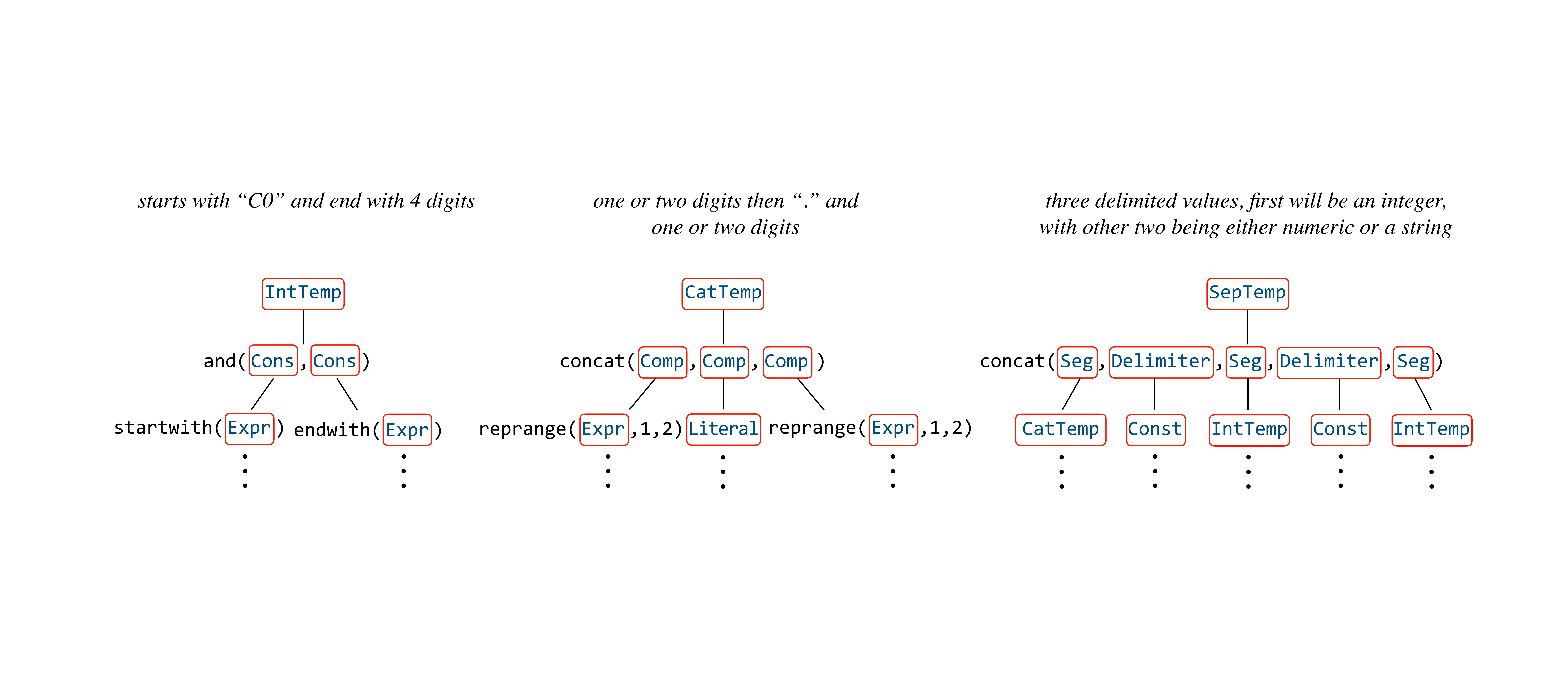}

  }

\begin{tabularx}{\linewidth}{l|X}
\toprule
\multirow{5}{*}{\regdsl{Cons}} & \regdsl{start[end]with(Expr) | not(start[end]with(Expr))}  \hfill \# must (not) start/end with\\
& \regdsl{contain(Expr) \gramdiv{} not(contain(Expr))} \hfill \# must (not) contain \\
& \regdsl{rep(<any>,k) \gramdiv{} repatleast(<any>,k) \gramdiv reprange(<any>,k,k) } \hfill \# length constraints \\
& \regdsl{AdvStartwithCons \gramdiv{} AdvEndwithCons} \hfill \# adversative macro (e.g., start with capitals except A) \\
& \regdsl{CondContainCons} \hfill \# conditional macro. (e.g. letter, if contained, must be after a digit) \\
\midrule
\multirow{2}{*}{\regdsl{Comp}} &  
\regdsl{Literal \gramdiv{} or(Literal,Literal,...)} \hfill \# literals like digits, letters, strings, or set of literals.\\
& \regdsl{rep(Expr,k) \gramdiv{} repatleast(Expr,k) \gramdiv{} reprange(Expr,k,k)\hfill }\# e.g, 3 digits, 2 - 5 letter, etc.\\
& \regdsl{optional(Comp) }\hfill \# components can be optional. \\
\bottomrule
\end{tabularx}
    \caption{Examples of our top-level templates and how they cover the three regexes in Figure~\ref{so_comp_reg}, and overview of sub-regexes (in table) that can possibly be derived from \regdsl{Cons} and \regdsl{Comp}. \regdsl{Expr} as a category here indicates various different constrained sets of sub-regexes. More detail about this structure is available in the full grammar in the appendix. }
    \label{fig:exs_rule}
\end{figure*}

\section{Structured Regex Generation Process}
\label{sec:grammar}
We first describe the structured generative process we adopt to produce the regexes in our dataset. For better readability, we denote regexes using a domain specific language (DSL) similar to regex DSLs in prior work \citep{deepregex, deepsketch}. Our DSL has the same expressiveness as a standard regular language and can be easily mapped back to standard regular expressions.\footnote{Refer to the appendix for details of our DSL.}

To collect the \turk{} dataset, \citet{deepregex} sampled regexes using a hand-crafted grammar similar to a standard regex DSL. However, regexes sampled from this process can easily have conflicts (e.g. \regdsl{and(<let>,<num>)}) or redundancies (e.g. \regdsl{or(<let>,<low>)}). One solution to this problem is rejection sampling, but this still does not yield regexes with compositional, real-world structure.

We show three prominent types of composition observed from \so{} in Figure~\ref{so_comp_reg}. Each regex above is built by assembling several sub-regexes together according to a high-level template: regex (a) is the intersection of two base regexes expressing constraints, regex (b) is a sequence of three simple parts, and regex (c) is a list of three segments delimited by a constant. We observe that these three templates actually capture a wide range of possible regex settings. The first, for example, handles password validation-esque settings where we have a series of constraints to apply to a single string. The second and third reflect matching sequences of fields, which may have shared structured (regex (c)) or be more or less independent (regex (b)).

\subsection{Structured Grammar}

\begin{figure}[t]
\centering
\includegraphics[width=\linewidth, trim=105 180 105 180,clip]{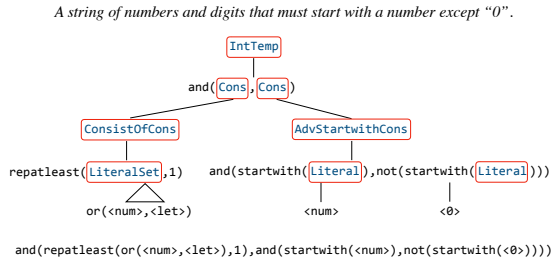}
\centering
\caption{The generation of a deep and complex regex using our grammar. Here, \regdsl{AdvStartwithCons} is a macro rule that yields a complex sub-tree with an adversative constraint.}
\label{fig:tree_exs}
\end{figure}

To generate realistic regexes in these forms, we rely on a structured hand-crafted grammar. The top level of our grammar specifies three templates distilled from \so{} examples: \textsc{Intersection}, \textsc{Concatenation}, and \textsc{Separation}, which mimic patterns of real-world regexes.  In Figure~\ref{fig:exs_rule}, we show how regexes in Figure~\ref{so_comp_reg} can be derived from our templates.
The \textsc{Intersection} template (left) intersects several base constraints with the \regdsl{and} operator; the \textsc{Concatenation} template (middle) concatenates several base components with the \regdsl{concat} operator.
\textsc{Separation} (right) is a more complex type, generating a list of constant-separated \textsc{Intersection} or \textsc{Concatenation} regexes which may be identical or share common components.

Across all templates, the components are sub-regexes falling into a few high-level types (notably \texttt{\small Cons} and \texttt{\small Comp}), which are depth-limited to control the overall complexity (discussed in Appendix~\ref{sec:impl_details}). To make these component regexes more realistic as well, we design several macro rules that expand to more than one operator. The macros are also extracted from real-world examples and capture complex relations like adversative (Figure~\ref{fig:tree_exs}) and conditional (Table~\ref{tab:quality}) relations.

Although our hand-crafted grammar does not cover every possible construction allowed by the regular expression language, it is still highly expressive. Based on manual analysis, our grammar covers 80\% of the real-world regexes in \so{}, whereas the grammar of \turk{} only covers 24\% (see Section~\ref{sec:dataset_analysis}). Note that some constructions apparently omitted by our grammar are equivalent to ones supported by our grammar: e.g., we don't allow a global \texttt{\small startwith} constraint in the \textsc{Concatenation} template, but this constraint can be expressed by having the first component of the concatenation incorporate the desired constraint.

\subsection{Sampling from the Regex Grammar}

Although our structural constraints on the grammar already give rise to more realistic regexes, we still want to impose further control over the generative process to mimic properties of real-world regexes. For example, there are sometimes repeating components in \textsc{Concatenation} regexes, such as regex (b) from Figure~\ref{so_comp_reg}.

We encourage such regexes by dynamically modifying the probability of applying the grammar rules while we are expanding a regex based on the status of the entire tree that has currently been induced. For example, suppose we are building regex (b) from Figure~\ref{so_comp_reg}, and suppose we currently have  \regdsl{concat(reprange(<num>, 1,2),concat(<.>,\textbf{Comp}))}, where \regdsl{\textbf{Comp}} is a non-terminal that needs to be expanded into a sub-regex. Because we already have \regdsl{reprrange(<num>,1,2)} and \regdsl{<.>} in the current tree, we increase the probability of expanding \regdsl{\textbf{Comp}} to generate these particular two sub-regexes, allowing the model to copy from what it has generated before.\footnote{This component reuse bears some similarity to an Adaptor Grammar \cite{AdaptorGrammar}. However, we modify the distributions in a way that violates exchangeability, making it not formally equivalent to one.}

In addition to copying, we also change the sampling distribution when sampling children of certain grammar constructs to control for complexity and encourage sampling of valid regexes. For example, the child of a \texttt{\small startwith} expression will typically be less complex and compositional than the child of a \regdsl{Comp} expression, so we tune the probabilities of sampling compositional AST operators like \regdsl{or} appropriately.


\section{Dataset Collection}

\subsection{Positive/Negative Example Generation}
The \so{} dataset \cite{deepsketch} shows that programmers often provide both positive and negative examples to fully convey their intents while specifying a complicated regex. Therefore, we augment our dataset with positive and negative examples for each regex. Our model will use these examples to resolve ambiguity present in the natural language descriptions. However, the examples can also help Turkers to better understand the regexes they are describing during the data collection process.


We aim to generate diverse and distinguishing examples similar to human-written ones, which often include corner cases that differentiate the ground truth regex from closely-related spurious ones. We can achieve this by enumerating examples that cover the states in the deterministic finite automaton (DFA) defined by the given regex\footnote{Recall that although our DSL is tree-structured, it is equivalent in power standard regexes, and hence our expressions can be mapped to DFAs.} and reject similar but incorrect regexes. We employ the Automaton Library \cite{automaton} to generate the examples in our work.
Positive examples are generated by stochastically traversing the DFA.

\begin{figure}[t]
\centering
\includegraphics[width=\linewidth, trim=160 250 155 250,clip]{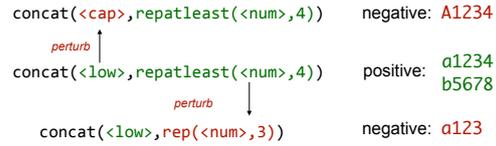}
\caption{The process of generating distinguishing negative examples by minorly perturbing each of the sub-regexes in the ground truth regex.}
\label{fig:gen_neg_exs}
\end{figure}

For negative examples, randomly sampling examples from the negation of a given regex will typically produce obviously wrong examples and not distinguishing negative examples as desired. Therefore, we propose an alternative approach shown in Figure~\ref{fig:gen_neg_exs} for generating negative examples. We apply minor perturbations to the ground truth regex to cause it to accept a set of strings that do not intersect with the set recognized by the original regex. The negative examples can be derived by sampling a positive string from one of these ``incorrect'' regexes.

For each regex in our dataset, we generate 6 positive examples and 6 negative examples. These numbers are comparable to the average number of examples provided by  \so{} users.

\subsection{Figure Generation}

\begin{figure}[t]
\centering
\includegraphics[width=\linewidth, trim=100 140 100 145,clip]{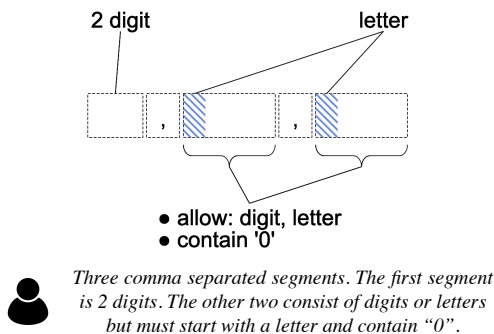}
\caption{An example automatically generated figure of a \textsc{Separation} regex and corresponding description annotated by a Turker.} \label{fig:fig_exs}
\end{figure}

As stated previously, we avoid the paradigm of asking users to paraphrase machine-generated regex descriptions, as this methodology can yield formulaic and artificial descriptions. Instead, we ask users to describe regexes based on figures that illustrate how the regex is built. We show one example figure of a \textsc{Separation} regex in Figure~\ref{fig:fig_exs}. In general, we abstract a given regex as a series of blocks linked with textual descriptions of its content and constraints. For instance, \regdsl{startwith} and \regdsl{endwith} are denoted by shading the head or tail of a block. By linking multiple blocks to shared textual descriptions, we hope to encourage Turkers to notice the correlation and write descriptions accordingly. Finally, we have different textual hints for the same concept: ``contain x'' in Figure~\ref{fig:fig_exs} may appear as ``have x'' elsewhere. These figures are rendered for each regex in the MTurk interface using JavaScript.

\subsection{Crowdsourcing}

\paragraph{Task} We collected the \streg{} dataset on Amazon Mechanical Turk (MTurk). For each HIT, the Turkers are presented with a regex figure and a set of positive/negative examples. Then, they are asked to write down several sentences describing the regex, as well as one additional positive example that matches the regex. We only accept a description if the submitted positive example is matched by the ground-truth regex; this helps filter out some cases where the Turker may have misunderstood the regex. We show an example HIT in Appendix~\ref{hit_prompt}.

In early pilot studies, we explored other ways of abstractly explaining regexes to Turkers, such as providing more examples and an associated set of keywords, yet none of these methods led to users generating sufficiently precise descriptions. By contrast, our figures fully specify the semantics of the regexes while only minimally biasing Turkers towards certain ways of describing them.

We generated 1,200 regexes (400 from each template), assigned each regex to three Turkers, and collected a total of 3,520 descriptions after rejecting HITs. In general, each Turker spent 2 to 3 minutes on each of the HITs, and we set the reward to be \$0.35. The total cost of collecting our dataset was \$1,512, and the average cost for each description is \$0.43.

\paragraph{Quality} To ensure the quality of collected responses, we require the Turkers to first take a qualification test which simply requires describing one regex that we have specified in advance. We then check that the description for this regex is sufficiently long and that it contains enough of our manually-written correct base regex concepts.

We manually observed from the responses that various styles were adopted by different Turkers for describing the same type of regexes. For instance, given regex (b) in Figure~\ref{so_comp_reg}, some Turkers tend to enumerate every component in order, describing it as \emph{one or two digits followed by a dot followed by one or two digits}; some other Turkers prefer grouping identical components and describing the components out of order, describing it as \emph{the first and third parts are one or two digits, and the second part is a dot}. These distinct styles lead to a diversity of linguistic phenomena, which is further analyzed in Section~\ref{sec:dataset_analysis}. Because we aim for high linguistic diversity in our dataset, we prohibited a single Turker from doing more than 300 HITs.

Furthermore, we found anecdotal evidence that the task was engaging for users, which we took as a positive signal for generation quality. We received messages about our HITs from some Turkers telling us that our HIT was ``really interesting'' and they ``enjoyed doing it.''


\begin{table}[t]
\small
\centering
\begin{tabularx}{\linewidth}{l|Y| Y| Y||Y}
\toprule
     Dataset & \textsc{KB13} & \textsc{Turk} & \textsc{StReg} &\textsc{SO}\\
\midrule
  size & 824 & 10000 & 3520 &62 \\
  \#. unique words & 207 & 557 & 873 & 301 \\
  avg. NL length & 8 & 12 & 33 & 25\\
  avg. reg size & 5 & 5 & 15 & 13\\
  avg. reg depth & 2.5 & 2.3 & 6.0 & 4.0\\
 \bottomrule
\end{tabularx}
\caption{Statistics of our dataset and prior datasets. Compared to \kb{} and \turk{}, our dataset contain more diverse language and more complex regexes, comparable to the real \so{} dataset. }
\label{tbl:datastats}
\end{table}

\begin{table*}[t]
    \centering
    \small
    \begin{tabularx}{\linewidth}{l|c|c|X}
    \toprule
        & \textsc{Turk} & \textsc{StReg} & Example NL from \textsc{StReg} \\
    \midrule
        
        multi-sentence  &0\% & 70\% & The string has 6 or more characters\textbf{.} The string must start with a digit.  \\

        ambiguity &2.3\% & 20.6\% & The sequence starts with a letter followed by 2 numbers. \\
        abstraction &0\% & 13.3\% & The first part of a single string consists of 1 or more ``0'' followed by 2 capital letters. The second part of the string must follow \textbf{the same rules}.\\
        non-local constraint & 0\% & 16.7\%  & There are 3 dash separated strings. The first is 1 to 4 ``A'' . The \textbf{second and third} consist of 1 or 2 ``x'' followed by 1 to 3 numbers and 2 letters. \\
        coreference & 5.1\% & 29.7\% &
        
        The string starts with a number. \textbf{It} ends with 1 to 4 lower or capital letters.\\
        \midrule
        condition relation & 0\% & 3.5\% & \textbf{If} there is a capital letter it must be after a digit.\\
        adversative  relation& 0\% & 3.7\% & The string start with capital letter \textbf{but} it should not be a ``A''. \\
    \bottomrule
    \end{tabularx}
    \caption{Qualitative analysis on 150 descriptions from \turk{} and our dataset (50 from each template). We show the percentage of examples containing each phenomenon. Our dataset features more of these challenging linguistic phenomena compared to prior synthetic datasets.}
    \label{tab:quality}
\end{table*}

\paragraph{Splitting the Dataset}
 Since our dataset consists of natural language descriptions written by annotators, there is possibly bias introduced by training and testing on the same annotators \cite{geva2019}. Therefore, in addition to the standard Train/Development/Test splits, we also form a Test-E (excluded) which consists only of annotations from annotators unseen in the training set. We ensure that Train, Dev, and both two test sets (Test and Test-E) have mutually exclusive regexes from each other (Test and Test-E can have common regexes), and Test-E is annotated entirely by a disjoint set of annotators from those who annotated the training or development set.
 The final size of the splits are: 2173 (61.7\%), 351 (10.0\%), 629 (17.9\%), 367 (10.4\%).

\section{Dataset  Analysis}
\label{sec:dataset_analysis}
We demonstrate the advantages of our dataset over prior datasets \cite{KB13,deepregex} through both quantitative and qualitative analysis.

We list the key statistics of our dataset as well as \kb{} and \turk{} for comparison in Table~\ref{tbl:datastats}. Compared to past synthetic datasets, our dataset has more diverse and sophisticated language. The average NL length of our dataset is twice as long as that of \turk{}, and the descriptions contain many more unique words even though our dataset contains fewer regexes.
In addition, our dataset contains more complex regexes that are closer to the complexity of real-world regexes found on StackOverflow, 
whereas  regexes in previous datasets are significantly simpler.

\paragraph{Manual Analysis} We further manually analyze 150 descriptions from past synthetic datasets and our dataset. Table~\ref{tab:quality} lists the proportion of descriptions containing each of several phenomena: examples that are \emph{multi-sentence}, examples with clear syntactic or semantic \emph{ambiguity}, examples using \emph{abstraction} to refer to different parts of the regex, examples invoking \emph{non-local constraints}, and examples with nontrivial \emph{coreference}. The language from our dataset is organic and diverse, since we allow Turkers to compose their own descriptions. We find that macros and complex constraints in our structured grammar can successfully trigger interesting language. For instance, the abstraction reflects repetition in concatenation regexes, and the bottom part of Table~\ref{tab:quality} reflects the complex macros.

\begin{table}[t]
    \centering
    \small
    \begin{tabularx}{\linewidth}{l|Y|Y|Y}
    \toprule
    & \kb{}&\textsc{Turk}&\textsc{StReg}\\
    \midrule
       Word Coverage  & 27.1\%& 34.4\% & \textbf{55.9\%}\\
       Regex Coverage &  23.5\%& 23.5\%& \textbf{84.3\%}\\
      \bottomrule
    \end{tabularx}
    \caption{Distribution mismatch analysis with respect to \so{} on past datasets and our dataset. Our dataset covers significantly more words and regexes, and is closer to the real-world dataset.}
    \label{tbl:coverage_so}
\end{table}

Furthermore, the complex and ambiguous language highlights the necessity of including examples together with language to fully specify a regex. For instance, ambiguity is common in our descriptions. However, many of the ambiguous descriptions can be resolved with the help of examples. Concretely, the description for \emph{ambiguity} from Table~\ref{tab:quality} can be easily interpreted as \regdsl{startwith(concat(<let>, repeat(<num>,2)))} while the ground truth is \regdsl{concat(<let>,repeat(<num>,2))}. By simply adding one negative example, ``a123'', the ground truth can be distinguished from the spurious regex. 

\paragraph{Comparison to \so{}} Since our goal was to produce realistic regex data, we analyze how well the real-world \so{} dataset is covered by data from \streg{} compared to other datasets \cite{KB13, deepregex}. We ignore 11 of the \so{} examples that involve the high-level \emph{decimal} concept, which is beyond the scope of our dataset and past synthetic datasets. In addition, we anonymize all the constants and integer parameters (e.g., \regdsl{repeat(<x>,9)} is anonymized as \regdsl{repeat(const,int)}). The statistics (Table~\ref{tbl:coverage_so}) suggest that our dataset is more highly similar to real-world regexes on StackOverflow, especially in terms of regex distribution.

\section{Methods}

We evaluate the accuracy of both existing grammar-based approaches and neural models, as well as a novel method that targets the multi-modal nature of our dataset.

\paragraph{Existing Approaches} \textsc{Semantic-Unify} \cite{KB13} is a grammar-based approach that relies on a probabilistic combinatory categorical grammar to build the regexes. \dpreg{} \cite{deepregex} directly translates natural language descriptions into regexes using a seq-to-seq model enhanced with attention \cite{attention} without considering examples. We re-implemented \dpreg{} with slightly different hyperparameters; we refer to our re-implementation as \textsc{DeepRegex (Ours)}. \dpfilter{} \cite{deepsketch} adapts \dpreg{} so as to take examples into account by simply filtering the $k$-best regexes based on whether a regex accepts all the positive examples and rejects all the negative ones.
 
\paragraph{Example-Guided Decoding}
Although \dpfilter{} is able to take  advantage of  positive and negative string examples, these examples are completely isolated in the training and inference phase. We propose to make use of examples during inference with the technique of over- and under- approximation \cite{lee} used in the program synthesis domain. The core idea of our approach is that, for each partially completed regex during decoding, we use the approximation technique to infer whether the regex can possibly match all positive or reject all negative examples. If this is impossible, we can prune this partial regex from our search. This approach allows us to more effectively explore the set of plausible regexes without increasing the computational budget or beam size.

As an example, consider the ground truth regex \regdsl{and(startwith(<low>),endwith(<num>))} with one corresponding positive example ``00x''. Suppose that the decoder has so far generated the incomplete regex \regdsl{and(startwith(<cap>),}. To produce a syntactically valid regex, the decoder needs to generate a second argument for the \regdsl{and}. By appending \regdsl{star(<any>)} as its second argument, we can see that there is no completion here that will accept the given positive example, allowing us to reject this regex from the beam. Under-approximation works analogously, completing regexes with maximally restrictive arguments and checking that negative examples are rejected.


We integrate the aforementioned technique in the beam decoding process by simply pruning out bad partial derivations at each timestep. We refer to this approach as \textsc{DeepRegex + Approx}.

\section{Experiments}
\subsection{Comparison to Prior Datasets}
\begin{table}[t]
\small
\centering
\begin{tabularx}{\linewidth}{l|Y Y Y}
\toprule
    Approach & \kb{} & \textsc{Turk} & \textsc{StReg} \\
  \midrule
\textsc{Semantic-Unify}   & 65.5\% & 38.6\% & 1.8\%\\
  \textsc{DeepRegex} (Locascio et al.) & 65.6\% & 58.2\% & $-$ \\
 \textsc{DeepRegex} (Ours) & 66.5\% & 60.2\% & 24.5\%\\
\midrule
  \textsc{DeepRegex + Filter} & 77.7\% & 83.8\% & 37.2\% \\
  \bottomrule
\end{tabularx}
\caption{DFA-equivalent accuracy on prior datasets and our dataset. The performance on our dataset using any model is much lower than the performance on existing datasets.}
\label{tbl:comp_results}
\end{table}
We evaluate the baseline models on \kb{}, \turk{}, and our dataset (Table~\ref{tbl:comp_results}).
The results show that our dataset is far more challenging compared to existing datasets. Traditional grammar baseline can scarcely solve our dataset.
The best baseline, \textsc{DeepRegex + Filter}, achieves more than 77.7\% on \kb{} and 83.8\% \turk{} when these datasets are augmented with examples, but can only tackle 37.2\% of our dataset. Additionally, the comparison between \dpreg{} and \textsc{Deepregex + Filter} demonstrates that simply filtering the outputs of neural model leads to a substantial performance boost on all the datasets. This supports the effectiveness of the way we specify regexes, i.e., using both natural language descriptions and examples.

\subsection{Detailed Results on \streg{}}
\begin{table*}[ht]
\small
\centering
\begin{tabular}{l|c c c c|c c c c}
\toprule
    \multirow{2}{*}{Approach} & \multicolumn{4}{c}{Test} & \multicolumn{4}{c}{Test-E} \\
    & Agg & Int & Cat & Sep & Agg & Int & Cat & Sep \\
  \midrule
    \textsc{Semantic-Unify}   & \phantom{0}2.1\% & \phantom{0}2.9\% & \phantom{0}3.1\% & \phantom{0}0.0\% & \phantom{0}1.4\% & \phantom{0}1.6\% & \phantom{0}2.4\% & \phantom{0}0.0\%\\
 \textsc{DeepRegex} (Ours) & 27.8\% & 20.7\% & 42.2\% & 19.2\% & 18.8\% & 18.0\% & 23.6\% &14.8\% \\
 \midrule
  \textsc{DeepRegex + Filter} & 42.6\% & 38.9\% & 55.2\% & 32.3\% & 28.1\% & 32.0\% & 32.5\% & 19.7\%\\
  \textsc{DeepRegex + Approx} & 48.2\% & 45.7\% \ & 59.6\% & 37.9\% & 36.0\% & 39.3\% & 40.7\% & 27.9\%\\ 
  \bottomrule
\end{tabular}
\caption{Results for models trained and tested on \streg{}. Using the examples (the latter two methods) gives a substantial accuracy boost, and \textsc{DeepRegex + Approx} is better than the post-hoc \textsc{Filter} method, but still only achieves 48.2\% accuracy on Test and 36.0\% on Test-E. Separation regexes are more difficult than the other two classes, and performance for all models drops on Test-E.}
\label{tbl:detail_result}
\end{table*}

Table~\ref{tbl:detail_result} shows the detailed accuracy regarding different regex templates on both Test and Test-E sets. Our \textsc{DeepRegex + Approx} achieves best accuracy with 5.6\% and 7.9\% improvement over \textsc{DeepRegex + Filter} on Test and Test-E, respectively, since it can leverage examples more effectively using over- and under- approximations during search.

Accuracy varies on different types of regexes. Generally, models perform the best on concatenation regexes, slightly worse on intersection regexes, and the worst on separation regexes. Concatenation regexes usually have straightforward descriptions in the form of listing simple components one by one. Intersection descriptions can be more complicated because of the high-level macros specified by our grammar. Separation descriptions are the most complex ones that often involve coreferences and non-local features. Performance on Test-E is 12\% lower than on Test for the models haven't been trained on patterns of the unseen annotators.

\begin{table}[t]
\small
    \centering
    \begin{tabularx}{\linewidth}{l|l|c|c|c}
        \toprule
        Train & Model& Acc & Equiv & Consistent \\
         Set & \dpreg{} && Found & Found  \\
        \midrule
          \textsc{Turk} & w/o Example & \phantom{0}0.0\% & \phantom{0}0.0\% & \phantom{0}7.8\% \\
         \midrule
         \textsc{StReg} & \textsc{+ Filter} & \phantom{0}9.8\% & \phantom{0}9.8\% & 21.6\% \\
        \textsc{StReg} & \textsc{+Approx}& 13.7\% & 17.6\% & 37.7\% \\
          \bottomrule
    \end{tabularx}
    \caption{The performance on \so{}-51 with models trained on \turk{} and our dataset. We report the fraction of examples where a DFA-equivalent regex is found (Acc), where a DFA-equivalent regex is found in the $k$-best list, and where a regex consistent with the examples appears in the $k$-best list. Models trained on \turk{} do not perform well in this setting, while our models can solve some examples.}
    \label{tab:trans_test}
\end{table}

\subsection{Transferability Results}

Finally, we investigate whether a model trained on our dataset can transfer to the \so{} dataset.
As in Section~\ref{sec:dataset_analysis}, we ignore instances requiring the decimal concept and only evaluate on the subset of \so{} with 51 instances.
We compare our dataset with \turk{} for this task. As shown in Table~\ref{tab:trans_test}, \dpreg{} trained on \turk{} completely fails on \so{} and even fails to predict reasonable regexes that are consistent with the examples. This is caused by the fact that the \turk{} dataset contains formulaic descriptions and shallow regexes that are not representative of real-world tasks. \dpreg{} trained on our dataset can at least achieve 9.8\% accuracy on \so{} dataset because the English descriptions in this dataset better match the desired task. Our \textsc{DeepRegex + Approx} model successfully solves 13.7\% and finds consistent regexes for 38\% of the tasks, which is credible given that the performance of the same model on Test-E set is only 30\%. Some additional challenges in \so{} are instances involving large numbers of constants or slightly more formal language since the SO users are mainly programmers. However, we believe the transfer results here show that improved performance on our dataset may transfer to \so{} as well, since some of the challenges also present in our Test-E set (e.g., unseen language).

\subsection{Human Performance Estimate}

It is difficult to hire Turkers to estimate a human performance upper bound, because our task requires reckoning with both the descriptions and positive/negative examples. Unlike many NLP tasks where an example with ambiguous language is fundamentally impossible, here the examples may actually still allow a human to determine the correct answer with enough sleuthing. But to perform this task, crowdworkers would minimally need to be trained to understand the DSL constructs and how they compose, which would require an extensive tutorial and qualification test. To do the task well, Turkers would need a tool to do on-the-fly execution of their proposed regexes on the provided examples.

We instead opted for a lighter-weight verification approach to estimate human performance. We adopted a post-editing approach on failure cases from our model, where we compared the model's output with the input description and examples and corrected inconsistencies. 

Specifically, we sample 100 failure examples from the test set (Test plus Test-E) and manually assess the failure cases. We find \textbf{78\%} of failure cases contain descriptions that describe all components of the target regexes, but our seq-to-seq models are insufficient to capture these.
There are truly some mis- or under-specified examples, such as not mentioning the optionality of one component or mistaking ``I'' for ``l'' in constants.
An additional \textbf{9\%} (out of 100) of the errors could be fixed using the provided examples. This leaves roughly 13\% of failure cases that are challenging to solve.

Considering that the model already achieves 43.6\% accuracy on the test set, we estimate human performance is around 90\%.\footnote{In addition, the first author manually wrote regexes for 100 randomly sampled examples and achieved an accuracy of 95\% (higher than the estimate). However, the author also has a strong prior over what synthetic regexes are likely to be in the data.}

\section{Related Work}

\paragraph{Data collection in semantic parsing} Collecting large-scale data for semantic parsing and related tasks is a long-standing challenge \cite{sempre, overnight}. \citet{overnight} proposed the generate-and-paraphrase framework, which has been adopted to collect datasets in various domains \cite{deepregex,howtosay,johnson2017clevr}. However, this process often biases annotators towards using formulaic language \cite{howtosay, herzig19}.

Similar to our work, past work has sought to elicit linguistically diverse data using visual elements for semantic parsing \cite{scone}, natural language generation \cite{novikova2016}, and visual reasoning \cite{nlvr, nlvr2}.
However, for these other tasks, the images used are depictions of an \emph{inherently graphical} underlying world state; e.g., the NLVR dataset \cite{nlvr} and NLVR2 \cite{nlvr2} are based on reasoning over the presented images, and the Tangrams dataset \cite{scone} involves describing shape transformations. By contrast, regexes are typically represented as source code; there is no standard graphical schema for depicting the patterns they recognize. This changes the properties of the generated descriptions, leading to higher levels of compositionality and ambiguity because what's being described is not naturally an image.

\paragraph{Program and regex synthesis} Recent research has tackled the problem of program synthesis from examples \cite{flashfill, gulwani2017programming, sygus,fidex,neo, robustfill,nye2019}. A closer line of work to ours uses both examples and natural language input \cite{sqlizer, deepsketch, andreas2018}, which involves fundamentally different techniques. However, our work does not rely on the same sort of program synthesizer to build final outputs \cite{sqlizer, deepsketch}. Moreover,  \citet{andreas2018} only use language at train time, whereas we use NL at both train and test time.

Finally, while several datasets on regex synthesis specifically have been released \cite{KB13, deepregex}, we are the first to incorporate examples in the dataset. 
Other methods have been proposed to parse natural language into regex via rule-based \cite{Ranta98}, grammar-based \cite{KB13}, or neural models \cite{deepregex, zexuan18, deepsketch}. Notably, \citet{zexuan18} also generate distinguishing examples to facilitate translation, but they require a trained model to generate examples, and we organically derive examples from the structure of regexes without additional input. 

\section{Conclusion}
We introduce \streg{}, a new dataset for regex synthesis from natural language and examples. Our dataset contains compositionally structured regexes paired with linguistically diverse language, and organically includes distinguishing examples. Better methods are needed to solve this dataset; we show that such methods might generalize well to real-world settings.

\section{Acknowledgments}

This work was partially supported by NSF Grant IIS-1814522, NSF Grant SHF-1762299, a gift from Arm, and an equipment grant from NVIDIA. The authors acknowledge the Texas Advanced Computing Center (TACC) at The University of Texas at Austin for providing HPC resources used to conduct this research. Thanks as well to the anonymous reviewers for their helpful comments.

\bibliography{acl2020}
\bibliographystyle{acl_natbib}

\newpage
\twocolumn
\appendix

\section{Regex DSL}
\begin{table}[h]
    \centering
\small
\begin{tabularx}{\linewidth}{l|X}
\toprule
Nonterminals \regdsl{r :=} &   \\
\quad \regdsl{startwith(r)} & \regdsl{r.*}\\
\gramdiv{} \quad \regdsl{endwith(r)} & \regdsl{.*r}  \\
\gramdiv{} \quad \regdsl{contain(r)} & \regdsl{.*r.*}  \\
\gramdiv{} \quad \regdsl{not(r)} & \regdsl{{\raise.17ex\hbox{$\scriptstyle\sim$}}r}  \\
\gramdiv{} \quad \regdsl{optional(r)} & \regdsl{r?}  \\
\gramdiv{} \quad \regdsl{star(r)} & \regdsl{r*}  \\
\gramdiv{} \quad \regdsl{concat($\regdsl{r}_1$, $\regdsl{r}_2$)} & \regdsl{$\regdsl{r}_1\regdsl{r}_2$}  \\
\gramdiv{} \quad \regdsl{and($\regdsl{r}_1$, $\regdsl{r}_2$)} & \regdsl{$\regdsl{r}_1\regdsl{\&}\regdsl{r}_2$}  \\
\gramdiv{} \quad \regdsl{or($\regdsl{r}_1$, $\regdsl{r}_2$)} & \regdsl{$\regdsl{r}_1\regdsl{|}\regdsl{r}_2$} \\
\gramdiv{} \quad \regdsl{rep($\regdsl{r}$,$k$)} & \regdsl{$\regdsl{r}\{k\}$} \\
\gramdiv{} \quad \regdsl{repatleast($\regdsl{r}$,$k$)} & \regdsl{$\regdsl{r}\{k,\}$} \\
\gramdiv{} \quad \regdsl{reprange($\regdsl{r}$,$k_1$,$k_2$)} & \regdsl{$\regdsl{r}\{k_1, k_2\}$} \\
\midrule
Terminals \regdsl{t :=} & \\
\quad \regdsl{<let>} & \regdsl{[A-Za-z]} \\
\gramdiv{} \quad \regdsl{<cap>} & \regdsl{[A-Z]}\\
\gramdiv{} \quad \regdsl{<low>} & \regdsl{[a-z]}\\
\gramdiv{} \quad \regdsl{<num>} & \regdsl{[0-9]}\\
\gramdiv{} \quad \regdsl{<any>} & \regdsl{.}\\
\gramdiv{} \quad \regdsl{<spec>} & \regdsl{[-,;.+:!@\#\_\$\%\&*=\^{}]}\\
\gramdiv{} \quad \regdsl{<null>} & \regdsl{$\emptyset$}\\
\bottomrule
\end{tabularx}
    \caption{Our regex DSL and the corresponding constructions in standard regular language. Our regex DSL is as expressive as and can be easily translated to standard regex syntax.}
    \label{tab:regex_dsl}
\end{table}
    
\normalsize


 \begin{figure*}[t]
  \small
     \centering
     \begin{tabularx}{\linewidth}{X}
        \toprule
        {\begin{tabularx}{\linewidth}{Y}
        \normalsize Intersection Template
        \end{tabularx}}\\
        \hline
        \gramrule{IntTemp}{Cons \gramdiv{} and(Cons,IntTemp)}\\
         \gramrule{Cons}{BasicCons \gramdiv{} LengthCons \gramdiv{} MacroCons}\\
         \gramrule{BasicCons}{not(BasicCons)}\\
          \gramrule{BasicCons} {startwith(ConsExpr)\gramdiv{}endwith(ConsExpr)\gramdiv{} contain(ConsExpr)} \\
         \gramrule{LengthCons}{rep(<any>,k)\gramdiv{} repatleast(<any>,k) \gramdiv{}reprange(<any>,k,k) }\\
         \gramrule{MacroCons}{ConsistOfCons\gramdiv{}AdvStartwithCons\gramdiv{} AdvEndwithCons \gramdiv{}
         CondContainCons  }\\
         \gramrule{ConsistOfCons}{repatleast(LiteralSet,1)}\\
         \gramrule{AdvStartwithCons}{and(startwith(Literal),not(startwith(Literal)))}\\
         \gramrule{AdvEndwithCons}{and(endwith(Literal),not(endwith(Literal)))} \\
        \gramrule{CondContainCons}{not(contain(concat(Literal,notcc(Literal))))}\\
        \gramrule{CondContainCons}{not(contain(concat(notcc(Literal),Literal)))}\\
        \gramrule{ConsExpr}{LiteralSet\gramdiv{}MinConsExpr\gramdiv{}concat(MinConsExpr,MinConsExpr)}\\
        \gramrule{MinConsExpr}{Literal\gramdiv{}rep(Literal,k)}\\
        \midrule
        {\begin{tabularx}{\linewidth}{Y}
        \normalsize Concatenation Template
        \end{tabularx}}\\
        \hline
        \gramrule{CatTemp}{Comp, concat(Comp, CatTemp)}\\
        \gramrule{Comp}{optional(Comp)}\\
        \gramrule{Comp}{BasicComp\gramdiv{} MacroComp}\\
        \gramrule{BasicComp}{CompExpr\gramdiv{}rep(CompExpr,k)\gramdiv{} repatleast(CompExpr,k) \gramdiv{}reprange(CompExpr,k,k)}\\
        \gramrule{MacroComp}{or(rep(<Literal>,k),rep(<Literal>,k))}\\
        \gramrule{MacroComp}{or(repatleast(<Literal>,k),repatleast(<Literal>,k))}\\
        \gramrule{MacroComp}{or(reprange(<Literal>,k,k),reprange(<Literal>,k,k))}\\
        \gramrule{CompExpr}{Literal\gramdiv{}LiteralSet}\\
        \midrule
        {\begin{tabularx}{\linewidth}{Y}
        \normalsize Separation Template
        \end{tabularx}}\\
        \hline
        \gramrule{SepTemp}{concat(Seg,Delimiter,Seg,Delimiter,Seg)}\\
        \gramrule{SepTemp}{concat(Seg,star(concat(Delimiter,Seg))}\\
        \gramrule{Seg}{IntTemp\gramdiv{}CatTemp}\\
        \gramrule{Delimiter}{CONST}\\
        \midrule
        {\begin{tabularx}{\linewidth}{Y}
        \normalsize Literals etc.
        \end{tabularx}}\\
        \hline
        \gramrule{Literal}{CC \gramdiv{} CONST \gramdiv{} STR} \quad \text{\# CONST can be any const character, STR can be any string values.}\\
        \gramrule{CC}{<num>\gramdiv{}<let>\gramdiv{}<low>\gramdiv{}<cap>\gramdiv{}<spec>}\\
        \gramrule{LiteralSet}{Literal\gramdiv{}or(Literal,LiteralSet)}\\
        \bottomrule
     \end{tabularx}
     \caption{Grammar rules for generating regexes in our dataset. Our grammar contains much more rules than a standard regex grammar, and is highly structured in that we have high-level templates and macros.
     }
     \label{fig:str_grammar}
 \end{figure*}
 \section{Details of Structured Grammar }

 \subsection{Grammar Rules}
 See Figure~\ref{fig:str_grammar}.
   \subsection{Implementation Details}

\label{sec:impl_details}
\paragraph{Intersection}
While building \textsc{Intersection} regexes, we impose context-dependent constraints mainly to avoid combinations of regexes that are redundant or in conflict. Conflicts often occur between a \regdsl{ComposedBy} constraint and the other constraints. A \regdsl{ComposedBy} constraint indicates the allowed characters; e.g., \regdsl{repeatatleast(or(<let>,<spec>),1)} means there can only be letters and special characters in the matched string. Therefore, 
when we already have such a constraint in the tree, 
we only allow the terminals to be selected from the valid subset of \regdsl{<let>} and \regdsl{<spec>}
while expanding the other subtrees.

This greatly reduce the chances of yielding empty regexes as well as redundant regexes (e.g., in \regdsl{and(repeatatleast(or(<let>,<spec>), 1),not(contain(<num>)))}, the second constraint is actually redundant).

\paragraph{Concatenation}
\textsc{Concatenation} regexes are a sequence of simple components. As stated above, our grammar encourages the phenomenon of repetition that commonly occurs in real regexes by copying existing sub-trees.

\paragraph{Separation}
\textsc{Separation} regexes have several subfields, which can be specified by either \textsc{Intersection} regexes or \textsc{Concatenation} regexes, and which are delimited by a constant. The fields of real regexes are often related, i.e., they share common components. For instance, the format of U.S. phone numbers is ``xxx-xxx-xxxx'' where ``x'' is a digit. Here the three fields are all digits but differ in length. Similar to the \textsc{Concatenation} template, we alter the distribution so as to copy the already generated subtrees.

We also allow a class of \textsc{Separation} with an arbitrary number of identical fields separated by a constant (e.g., \emph{a list of comma-separated numbers}).


\paragraph{Complexity Control}
We aim to create a collection of complicated regexes, but we do not wish to make them needlessly complex along unrealistic axes. We assess the complexity of generated regexes using a measure we call \emph{semantic complexity}, which roughly measures how many factors would need to be specified by a user. Generally, each constraint or components counts for one degree of semantic complexity, e.g., \regdsl{not(contain(x))} and \regdsl{repeat(x,4)} are of complexity level one. High-level macro constraints are of complexity level two since they need more verbal explanation. We limit the complexity degrees all of our generated regexes to be strictly no more than six. More details about the number of nodes and depth of our regexes can be found in Section~\ref{sec:dataset_analysis}.

\section{HIT Example}
\label{hit_prompt}

\begin{figure*}[h]
\begin{tabularx}{\linewidth}{X}
\hline
\vspace{3mm}
\textbf{Instructions:}\\
In this task, you will be writing down descriptions of the patterns you see in a group of strings. For each HIT, you'll be given a figure visually specifying a pattern and a few examples of strings following or not following the pattern to help you to understand it. Please write a description (generally 1-4 sentences) that describes the pattern. In addition, please write one additional string that follows the pattern.\\
Things to keep in mind:\\
 $\bullet$ Please describe the pattern underlying the string examples, not the sequence of strings itself. Do not write things like ``the first line ..., the second line ....'' \\
 $\bullet$ Try to be precise about describing the pattern, but also concise. Don't describe the same property of the strings in multiple ways.\\
 $\bullet$ You are not required to use the keywords in the figure. If you can think of another way to express the intent, that's okay.\\
 $\bullet$ Please try to write natural and fluent sentences.\\
 $\bullet$ Additional string example must be different.\\

 \includegraphics[width=0.5\linewidth, trim=100 100 80 100,clip]{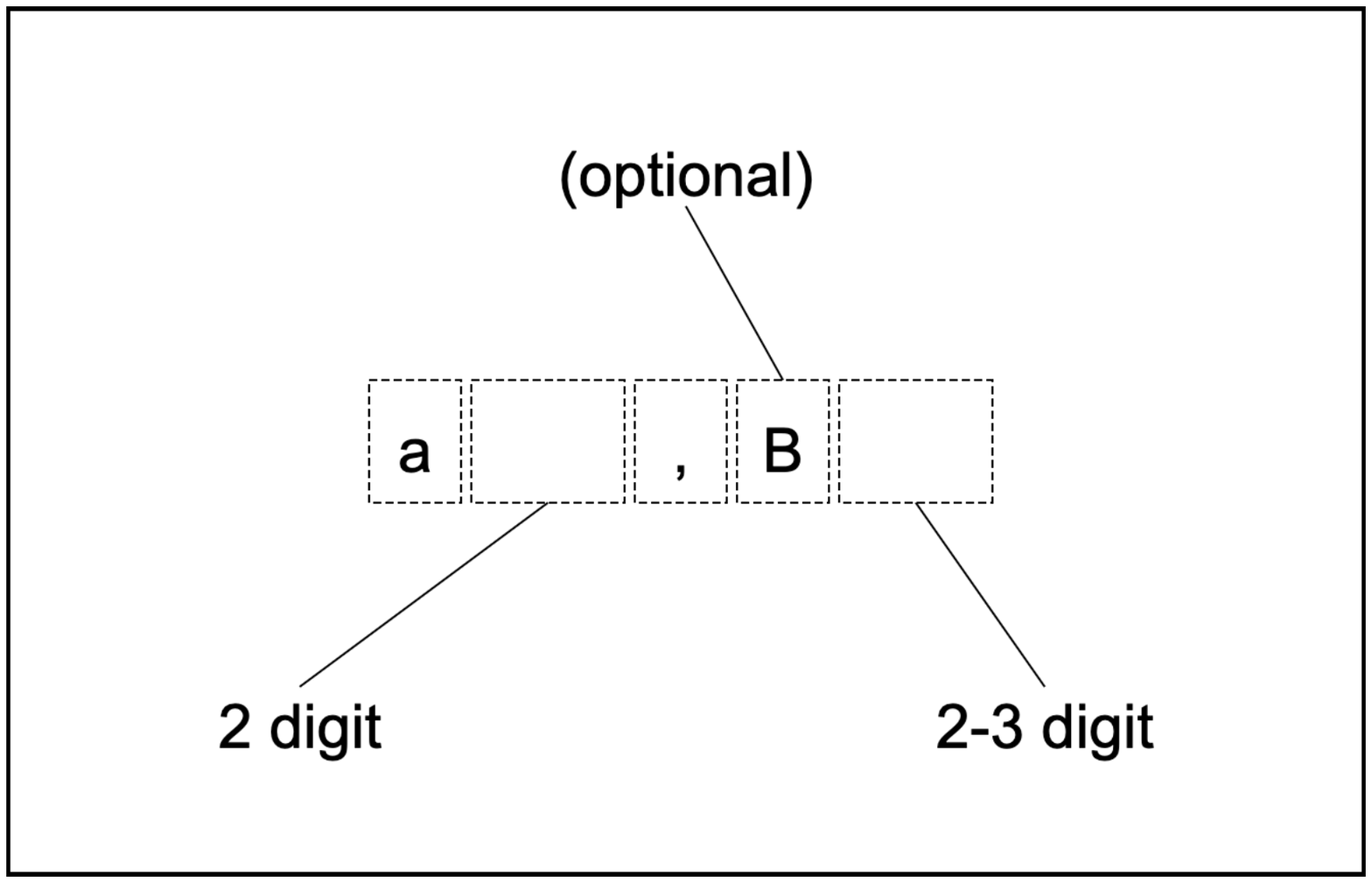}\\

\textbf{Example strings that follow the pattern:}\\
a51,B457\\
a74,B23\\
a09,849\\

\textbf{Example strings that do not follow the pattern:}\\
b55,B193\\
a7,B23\\
a09,1\\
\vspace{3mm}\\
\hline

\end{tabularx}

\vspace{8mm}

    \caption{HIT prompt for the description writing task. We particularly emphasize in the instructions that Turkers should use precise and original language.}
    \label{fig:hit_exs}
\end{figure*}

See Figure~\ref{fig:hit_exs}.

\end{document}